\documentclass[10pt,journal,compsoc]{IEEEtran}
\ifCLASSOPTIONcompsoc
  \usepackage[nocompress]{cite}
\else
  \usepackage{cite}
\fi

\usepackage[utf8]{inputenc} % allow utf-8 input
\usepackage[T1]{fontenc} % use 8-bit T1 fonts
\usepackage{hyperref}  % hyperlinks
\usepackage{url}   % simple URL typesetting
\usepackage{booktabs}  % professional-quality tables
\usepackage{amsfonts}  % blackboard math symbols
\usepackage{nicefrac}  % compact symbols for 1/2, etc.
\usepackage{microtype}  % microtypography
\usepackage{xcolor}   % colors
\usepackage{amsmath}
\usepackage{amssymb}
\usepackage{graphicx}
\usepackage{booktabs}
\usepackage{wrapfig}
\usepackage{lipsum}
\usepackage{microtype}
\usepackage{stmaryrd}
\usepackage{xspace}
\usepackage{floatrow}
\usepackage{amsthm}
\usepackage{multirow}
\usepackage{caption}

\newtheorem{assumption}{Assumption}

\newtheorem{lemma}{Lemma}
\newtheorem{theorem}{Theorem}

\newcommand{\scal}[2]{<{#1},{#2}>}

\newcommand{\norms}[1]{||{#1}||^2}
\newcommand{\esp}{\mathbb{E}}

\begin{document}
\title{Improved optimization strategies for deep Multi-Task Networks}

\author{Lucas~Pascal,
        Pietro~Michiardi,
        Xavier~Bost,
        Benoit~Huet,
        and~Maria~A. ~Zuluaga}% <-this % stops a space
% \IEEEcompsocitemizethanks{
% \IEEEcompsocthanksitem L. Pascal ...\protect\\
% % note need leading \protect in front of \\ to get a newline within \thanks as
% % \\ is fragile and will error, could use \hfil\break instead.
% E-mail: lucas.pascal@eurecom.fr
% % \IEEEcompsocthanksitem P. Michiardi ...\protect\\
% % E-mail: lucas.pascal@eurecom.fr
% % \IEEEcompsocthanksitem P. Michiardi ...\protect\\
% % E-mail: lucas.pascal@eurecom.fr
% % \IEEEcompsocthanksitem P. Michiardi ...\protect\\
% % E-mail: lucas.pascal@eurecom.fr
% % \IEEEcompsocthanksitem P. Michiardi ...\protect\\
% % E-mail: lucas.pascal@eurecom.fr
% \thanks{Manuscript received April 19, 2005; revised August 26, 2015.}}

% The paper headers
% \markboth{Journal of \LaTeX\ Class Files,~Vol.~14, No.~8, August~2015}%
% {Shell \MakeLowercase{\textit{et al.}}: Bare Demo of IEEEtran.cls for Computer Society Journals}

\IEEEtitleabstractindextext{%
\begin{abstract}
In Multi-Task Learning (MTL), it is a common practice to train multi-task networks by optimizing an objective function, which is a weighted average of the task-specific objective functions. Although the computational advantages of this strategy are clear, the complexity of the resulting loss landscape has not been studied in the literature. Arguably, its optimization may be more difficult than a separate optimization of the constituting task-specific objectives. In this work, we investigate the benefits of such an alternative, by alternating independent gradient descent steps on the different task-specific objective functions and we formulate a novel way to combine this approach with state-of-the-art optimizers. As the separation of task-specific objectives comes at the cost of increased computational time, we propose a random task grouping as a trade-off between better optimization and computational efficiency. Experimental results over three well-known visual MTL datasets show better overall absolute performance on losses and standard metrics compared to an averaged objective function and other state-of-the-art MTL methods. In particular, our method shows the most benefits when dealing with tasks of different nature and it enables a wider exploration of the shared parameter space. We also show that our random grouping strategy allows to trade-off between these benefits and computational efficiency.
\end{abstract}

% Note that keywords are not normally used for peerreview papers.
}

% make the title area
\maketitle

\IEEEdisplaynontitleabstractindextext

\IEEEpeerreviewmaketitle

\IEEEraisesectionheading{\section{Introduction}}\label{sec:introduction}
% \section{Introduction}\label{sec:introduction}
Multi-Task Learning (MTL) consists in jointly, rather 
than individually, learning multiple tasks with the goal of improving generalization performance. This is done by training a shared model for all tasks~\cite{caruana_multitask_1997}. In deep MTL, the shared model consists in the parameters of a deep network.

The optimization objective in MTL is formulated as the combination into a single aggregated objective function, in the form of a weighted sum of all the task-specific objective functions~\cite{caruana_multitask_1997}, which the shared model is trained to minimize. This formulation is particularly convenient for deep MTL. First, it is computationally efficient, since a single gradient descent step (i.e. one forward and one backward propagation)
optimizes the shared parameters  w.r.t. every task, with the only extra computational cost coming from the task-specific parts of the network. Second, in the case of convex task-specific objective functions, it guarantees convergence to a minima, since the average objective function remains convex.

Despite its simplicity, computational efficiency and wide adoption~\cite{gao_nddr-cnn_2019,he_deep_2016-1,kokkinos_ubernet_2017-1,liu_end--end_2019-2,lu_fully-adaptive_2017-1,misra_cross-stitch_2016-1,mordan_revisiting_2018-1}, the use of such an objective function can be problematic due to the complex loss landscapes of deep networks. %, which are not convex in the general case. 
In fact, the landscape complexity increases exponentially with the depth of the network \cite{li_visualizing_2018}. As a result, the network parameter update can be sub-optimal when the task gradients conflict, or when one task dominates because of a much higher gradient magnitude  w.r.t. the other tasks~\cite{Vandenhende2021}. This has led researchers to investigate alternative methods to balance the task-specific objectives using adaptable task-specific weights \cite{chen_gradnorm_2018,chennupati_multinet_2019,guo_dynamic_2018,kendall_multi-task_2018-1,liu_end--end_2019-2} or by altering the gradient of the objective function \cite{chen_just_2020,yu_gradient_2020} in cases of conflicts between different tasks. 

Partitioning methods~\cite{bragman_stochastic_2019,maninis_attentive_2019-1,pascal_maximum_2020, strezoski_many_2019-1} are a recent set of works that propose to circumvent the problems associated to the optimization of MTL's aggregated objective function by using task-specific partitioning of the parameters. 
They aim to reduce the occurrence of conflicts between tasks by letting each task adapt the shared representation of the network with a few task-specific weights.
% They aim to reduce the occurrence of conflicts between tasks and give more flexibility to the network.{\color{red} What do you mean by flexibility?} 
By construction, these methods optimize each task-specific objective function \textit{alternately and independently}, which is not equivalent to optimizing the average sum of the different task-specific objectives, i.e. the aggregated loss. Despite promising results, the difference between the two schemes has never been studied, nor are there theoretical guarantees that this strategy can lead to similar results as the aggregated loss one. This is in part can be explained by the fact that the optimization scheme of partitioning methods is always coupled with parameter partitioning, making it difficult to evaluate which part of the contribution is due to the partitioning, and which one is due to the optimization scheme.

%Some more recent works try to circumvent these optimization difficulties by using task-specific partitionings of the parameters, in order to allow different uses of the shared parameters for each task.
%They aim at reducing the happenings of conflicts between tasks, and giving more flexibility to the network. 
%These "partitioning" methods, by construction, optimize each task-specific objective function alternately and independently, which is not equivalent to optimizing an average sum of the different task-specific objective functions. However this difference is never studied, and the optimization scheme in these works is always coupled with parameter partitioning, making it difficult to evaluate which part of the contribution is due to the partitioning, and which is due to the optimization scheme.

Motivated by the promising results from recent works~\cite{bragman_stochastic_2019,maninis_attentive_2019-1,pascal_maximum_2020,strezoski_many_2019-1}, in this paper we propose to study the task-specific alternate and independent optimization used in partitioning methods and we formulate improvements to it by making the following contributions. 1) 
We prove that this optimization strategy provides convergence guarantees similar to the aggregated loss optimization for stochastic gradient descent in the convex case, and we provide the associated convergence bound.
2) We show that current alternated optimization schemes do not operate truly independent task-specific update steps, due to the momentum mechanisms of most of the existing optimizers, which memorize previous updates, thus bringing them into the current learning steps. We therefore propose a novel alternated  optimization scheme that performs truly independent task-specific update steps with momentum-based state-of-the-art optimizers. 
3) To account for the losses in computational efficiency that alternate independent updates incur into, we introduce a task grouping strategy to reduce training time when dealing with a high number of tasks. Using three well-known benchmark datasets, we validate our work and highlight our contributions by demonstrating that the proposed optimization strategies consistently lead to substantial improvements in terms of both generalization performance and benchmark results over existing state-of-the-art methods.
\section{Related work}\label{sec:related}

Our work focuses on the study of the alternate and independent optimization of task-specific objective functions used in partitioning methods, which we review in the following. As most MTL methods rely on the use of an aggregated objective function, we provide a brief overview of such approaches. We refer the reader to \cite{crawshaw_multi-task_2020, Vandenhende2021} for a broader review of this family of methods. 
% In this section we briefly describe existing MTL optimization methods, under the scope of this paper, which is the opposition between the widely used aggregated objective function optimization and the independent optimization of task-specific objective functions used in partitioning methods. A more detailed and complete description of existing MTL methods can be found in . 

\textbf{Aggregated loss.}
Most MTL works optimize an aggregated objective function containing all the task-specific objective functions~\cite{caruana_multitask_1997} and they develop different strategies to mitigate the problems associated to the complexity of the loss.
A first family of methods uses dynamic loss weighting strategies to control the influence of tasks in the main objective function and to account for the learning dynamics of the different tasks~\cite{chen_gradnorm_2018, guo_dynamic_2018, kendall_multi-task_2018-1, liu_end--end_2019-2}.
Other works formulate the problem as a multi-objective optimization task~\cite{desideri_multiple-gradient_2012}, which converges to a Pareto optimal solution, from which no task can be improved without hurting another one~\cite{lin_pareto_2019,sener_multi-task_2018-1}. This greedy strategy can be particularly sensitive to strongly interfering tasks, quickly converging into poor minima~\cite{pascal_maximum_2020}.
A more recent line of works proposes to modify the task-specific gradients before averaging them, when there are conflicting gradient directions, under the hypothesis that they can be destructive for some tasks~\cite{chen_just_2020,yu_gradient_2020}. 
Although some of these works provide a better convergence in specific conflicting regions of the loss landscape, the guarantees are very local and become of relative importance in the case of complex non-convex loss landscapes, for which a wider exploration of the parameter space might be preferred over greedy strategies. From this perspective, in this work we favor the optimization strategies adopted in partitioning methods, as they do not make any assumptions that are detrimental to learning and they avoid greedy improvements.

\textbf{Partitioning methods.}
These works define task-specific parameter partitioning, to allow tasks specify their own usage of the shared parameters, while potentially reducing cases of task interference~\cite{bragman_stochastic_2019,maninis_attentive_2019-1,pascal_maximum_2020, strezoski_many_2019-1}. They \textit{alternately} compute \textit{independent} forward and backward passes for each task according to the corresponding task-specific partitioning, and alternate these optimization steps among the different tasks. The partitioning itself can then be of different kinds. In \cite{maninis_attentive_2019-1}, the authors use task-specific Squeeze and Excitation blocks to create a real-valued partitioning. Other works apply a binary partitioning to the parameters, either by training the partition ~\cite{bragman_stochastic_2019} with the Gumbel Softmax trick \cite{jang_categorical_2017,maddison_concrete_2017}, by fixing static partitions~\cite{strezoski_many_2019-1} or by making the partitioning randomly evolve~\cite{pascal_maximum_2020}. Similar strategies are used in \cite{Mallya_2018,mancini_adding_2018}, without involving any shared weights optimization, since the partitioning is applied to a frozen pre-trained network.
Regarding the optimization strategy,
% {\color{red} From here: Mention the thing of the optimization as it is what we care about.}
only \cite{maninis_attentive_2019-1} discusses it explicitly as a way to exploit the benefits of partitioning. Therefore, it has not been studied in isolation in any previous work.
% These works however never study the separation of the task-specific objective functions alone, and instead combine it with parameters partitioning. 
Moreover, these works do not consider that moving average mechanisms (e.g. momentum) included in state-of-the-art optimizers can mix previous gradient descent directions from different tasks.
% {\color{red}memories of past parameter updates from differents tasks NOT CLEAR what you mean by memories}. 
This means that the individual objective functions are only partially separated. Our work proposes a full separation of the individual objective functions, by defining individual moving average mechanisms for every task. To account for the incurred computational overhead of such methods, we propose a random task grouping as a trade-off between the benefits of alternate and independent task-specific optimization and computational efficiency.

\section{Alternate and independent optimization of task-specific objective functions}\label{sec:method}
We first formalize the standard MTL optimization setup using an aggregated loss before our contributions. 
% We first formalize the standard MTL optimization setup using an aggregated loss, that we denote as MTL-SUS. We then present two different strategies for alternate and independent optimization of the task-specific objective functions, MTL-IUS and MTL-IO. We finally present a random task grouping strategy which aims at mitigating the computational overhead of these strategies.

\subsection{Standard MTL optimization with aggregated loss}
Let $\xi_t$ be an input data instance randomly sampled during optimization step $t$, and $k$ the index of the $N$ tasks. We define the aggregated multi-task objective function to minimize as 
\begin{eqnarray}\label{eq:std_loss}
F(w_t, \xi_t) = \sum_{k=1}^N c^{(k)}\cdot F^{(k)}(w_t, \xi_t),
\end{eqnarray}
where $F^{(k)}$ is the objective function associated to task $k$ using shared parameters $w_t$, and $c^{(k)}$ are the task-specific weighting coefficients. For the sake of simplicity in the notation, but without loss of generality, we consider here a uniform weighting of the different task objective functions, i.e. $c^{(k)}=1$. Hereinafter, we use superscripts to index elements associated to different tasks, and subscripts to index the steps of the optimization process.

When using stochastic gradient descent (SGD) to optimize Eq.~\ref{eq:std_loss}, the shared parameters $w$ at step $t+1$ are updated according to the following rule:
\begin{align}
\label{eq:sgd_update}
w_{t+1} = w_t - \eta_t \sum_{k=1}^N \frac{\partial}{\partial w_t}F^{(k)}\left(w_t, \xi_t\right),
\end{align}
%
% \begin{align}\label{eq:sgd_update}
% w^{(k+1)} = w^{(k)} - \eta \sum_{i=1}^T \frac{\partial}{\partialw^{(k)}} \mathcal{L}_i\left(f\left(x,w^{(k)}\right),y_i\right)
% \end{align}
%
where $\eta_t$ is the learning rate. 
For the ease in notation, let us denote $g^{(k)}(w_t, \xi_t)$ the derivative of $F^{(k)}$  w.r.t. the parameters $w_t$. Eq.~\ref{eq:sgd_update} can be rewritten as
% \begin{align}
% w^{(k+1)} = w^{(k)} - \eta \sum_{i=1}^T g_i\left(w^{(k)}\right),
% \label{eq:MTL_update}
% \end{align}
\begin{align}\label{eq:sgd_g_update}
w_{t+1} = w_t - \eta_t \sum_{k=1}^N g^{(k)}\left(w_t, \xi_t\right).
\end{align}
In the remaining of the document, we will refer to this optimization strategy with an aggregated loss as MTL with Shared Update Steps (MTL-SUS) to reflect the characteristics of its update rule (Eq.~\ref{eq:sgd_g_update}).

\subsection{Alternate and independent optimization of task-specific objective functions for SGD}
\label{subsec:IUS_SGD}
We now formalize the SGD update step for the alternate and independent optimization scheme used in state-of-the-art MTL partitioning works~\cite{bragman_stochastic_2019,maninis_attentive_2019-1,pascal_maximum_2020,strezoski_many_2019-1} and we provide convergence bounds in the special case of convex objective functions.

%\subsubsection{MTL Individual Update Steps (MTL-IUS)}
We define a multi-task update step as the alternate Individual Update Steps (IUS) of each of the $N$ task-specific objective functions, and denote $\smash{w_t^{(k)}}$ the parameters optimized  w.r.t. task $k$, during the $t$-th update step.
The individual update step $t+1$ of task $k$ is:
\begin{equation}
 w_{t+1}^{(k)} =
 \begin{cases}
  w_{t}^{(N)} - \eta_t \cdot g^{(k)}\left(w_t^{(N)}, \xi_t\right), \quad k=1
  \\[10pt]
 w_{t+1}^{(k-1)} - \eta_t \cdot g^{(k)}\left(w_{t+1}^{(k-1)}, \xi_t\right), \quad \forall \quad k > 1. 
 \end{cases}  
 \label{eq:IUS_SGD}
\end{equation}
To reflect the properties of the update rule (Eq.~\ref{eq:IUS_SGD}), we denote this optimization strategy MTL-IUS.

% \subsubsection{Convergence in the convex case}
MTL holds analogies with the federated learning problem~\cite{konecny2015,li2020convergence, shokri2015}, where each involved device accounts for one task objective function, with communication after each update step. MTL-SUS corresponds to a full-device participation, and MTL-IUS to a single device participation.
Using this analogy, we take inspiration from~\cite{li2020convergence} to prove that alternate and independent optimization schemes using the MTL-IUS rule converge to the optimum in the case of convex objective functions.
% Considering that our formulation is very close to that of a federated learning problem with partial device participation, we take inspiration from~\cite{li2020convergence} to prove that MTL-IUS converges to the optimum.
% We first state usual assumptions for stochastic gradient descent, define some notations, and finally provide the convergence bound of MTL-IUS. }
% {\color{red}We study here the convergence of MTL-IUS in the simple case where all the different tasks objective functions are convex. Considering that our setting can be seen as a particular case of Federated Learning \cite{konecny2015, shokri2015, li2020convergence} with partial device participation, we essentially adapt this part from \cite{li2020convergence}. We first state usual assumptions for stochastic gradient descent, define some notations, and finally provide the convergence bound of MTL-IUS. }
\begin{assumption}
\label{ass:smooth}
$F^{(1)},...,F^{(N)}$ are all $L$-smooth: for all $v$, $w$, $F^{(k)}(v) \leq F^{(k)}(w) + (v-w)^T \nabla F^{(k)}(w) + \frac{L}{2} ||v-w||_2^2$.
\end{assumption}

\begin{assumption}
\label{ass:convex}
$F^{(1)},...,F^{(N)}$ are all $\mu$-strongly convex: for all $v$, $w$, $F^{(k)}(v) \geq F^{(k)}(w) + (v-w)^T \nabla F^{(k)}(w) + \frac{\mu}{2} ||v-w||_2^2$.
\end{assumption}

\begin{assumption}
\label{ass:stochastic_var}
Let $\xi_t$ be uniformly sampled from the data. The variance of stochastic gradients is uniformly bounded: $\mathbb{E}||\nabla F^{(k)}(w_t, \xi_t) - \nabla F^{(k)}(w_t)||^2 \leq {\sigma^{(k)}}^2$.
\end{assumption} 

\begin{assumption}
\label{ass:stochastic_norm}
The expected squared norm of stochastic gradients is uniformly bounded: $\mathbb{E}||\nabla F^{(k)}(w_t, \xi_t)||^2 \leq G^2$.
\end{assumption}

Let us denote $F^*$ and ${F^{(k)}}^*$ as the minima values of $F$ and $F^{(k)}$ respectively, and $w^*$ as the minimizer of $F$. We thus define
\begin{equation*}
 \Gamma = F^* - \frac{1}{N}\sum_{k=1}^N {F^{(k)}}^*,
\end{equation*}
which can be considered as the heterogeneity of the different task-specific objective functions.

\begin{theorem}
\label{theorem:convergence}
Let $\gamma \geq 2\frac{L}{\mu}-1$ and $\eta_t = \frac{2}{\mu(\gamma+t)}$ the learning rate. Under assumptions \ref{ass:smooth}, \ref{ass:convex}, \ref{ass:stochastic_var} and \ref{ass:stochastic_norm}, the optimization scheme MTL-IUS presented in Eq.~(\ref{eq:IUS_SGD}) satisfies
\begin{equation}
 \label{eq:convergence}
 \mathbb{E}[F(w_T)] - F^* \leq \frac{L}{\gamma + T} \left( \frac{2B}{\mu^2} + \frac{(\gamma+1)}{2} \mathbb{E}||w_1 - w^*||^2 \right),
\end{equation}
with $B = \sum_{k=1}^N p_k^2 \sigma_k^2 + 2L\Gamma + G^2$. 
\end{theorem}

\textbf{Proof:}
See appendix \ref{app:proof} $\qedsymbol$

Theorem~\ref{theorem:convergence} guarantees the convergence of MTL-IUS to the optimum in $\mathcal{O}(\frac{1}{T})$ with a decreasing learning rate for stochastic gradient descent. However, it should be noted that the bound does not increase directly with the number of tasks, but with the heterogeneity term $\Gamma$, which corresponds to the average deviation between the optimum $F^*$ and the values of the task-specific objective functions $\smash{{F^{(k)}}^*}$ at that point. 

\subsection{Alternate and independent optimization of task-specific objective functions for moving-average based optimizers}

In most of deep learning's literature and in current practice, more complex optimizers than a regular SGD are preferred. State-of-the-art optimizers, such as Adam~\cite{kingma_adam_2017-1}, generally use exponential moving average mechanisms, e.g. momentum, to smooth and regulate the optimization trajectory. These mechanisms can partially compromise the independence of the task-specific update steps, i.e. MTL-IUS, since every learning step includes descent directions of previous tasks.
In this section, we propose a new update rule that allows to use Individual Optimizers (IO), providing improved optimization independence to the different tasks.

\textbf{MTL-IUS with momentum.} We first reformulate the update step of MTL-IUS with a momentum-based optimizer,
% We first formalize the {\color{red} direct adaptation\textbf{ what do you mean by adaptation}} of MTL-IUS to moment-based optimizers, 
as it is done in partitioning MTL works~\cite{bragman_stochastic_2019,maninis_attentive_2019-1,pascal_maximum_2020}.

\begin{figure*}[t]
 \centering
 \includegraphics[width=0.49\linewidth]{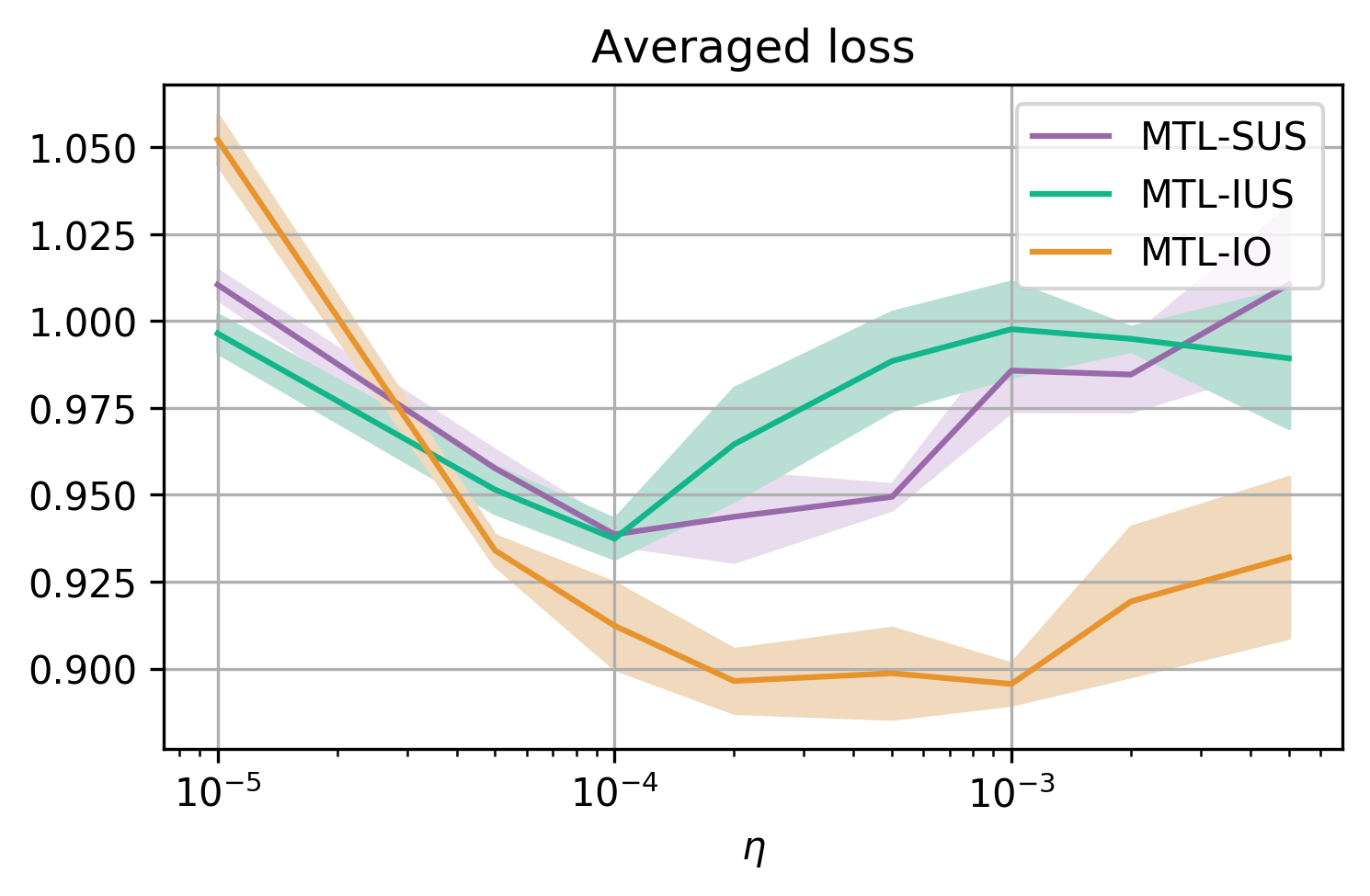}
 \includegraphics[width=0.49\linewidth]{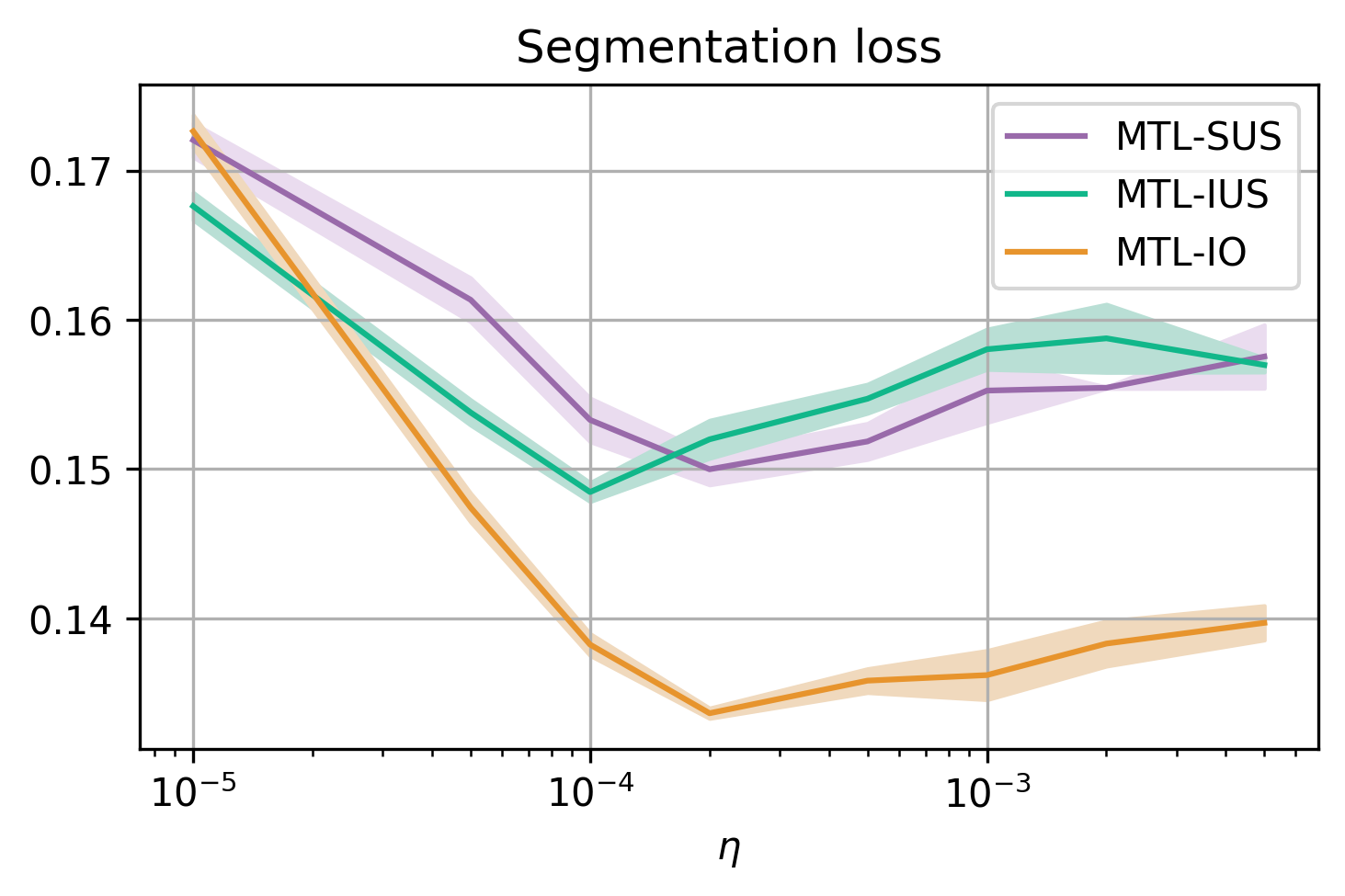}
 \includegraphics[width=0.49\linewidth]{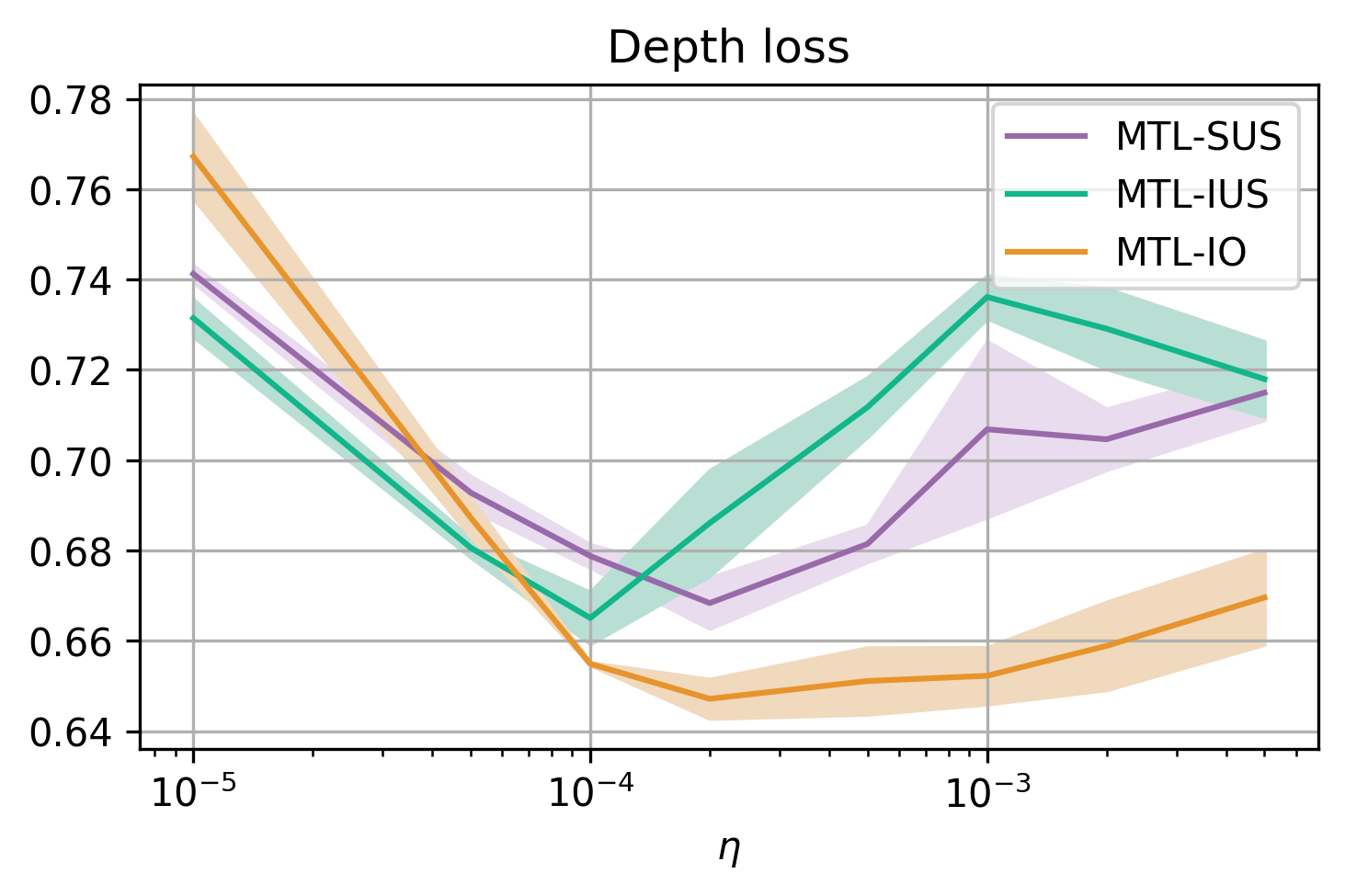}
 \includegraphics[width=0.49\linewidth]{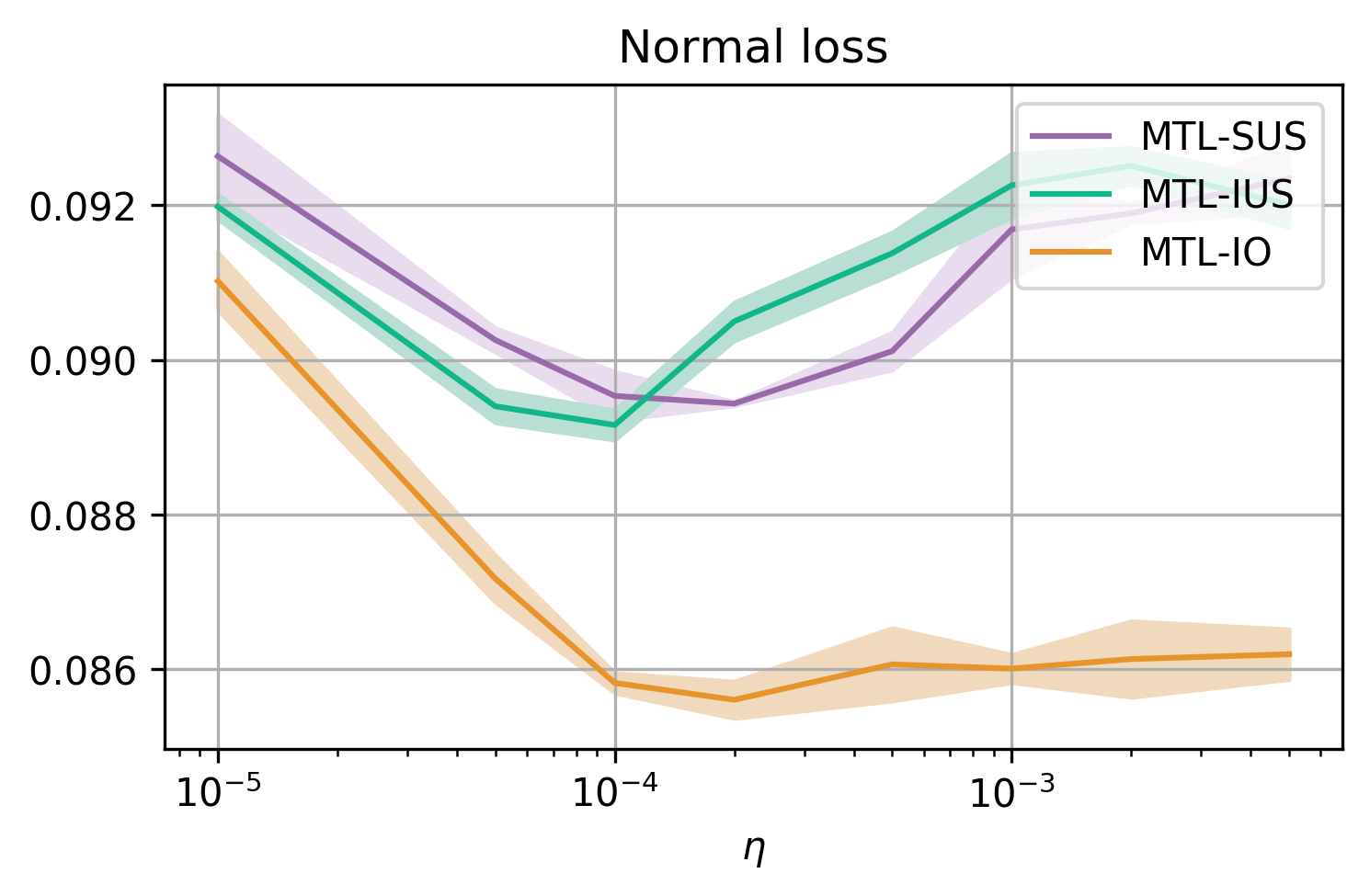}
 \caption{NYUv2 best averaged and task-specific validation losses for the different tasks with respect to the learning rate $\eta$.}
 \label{fig:nyu_loss}
\end{figure*}

%\subsubsection{MTL-IUS with momentum}
Let us define $\hat{m}$ a generic optimizer rule, which adjusts the gradient descent direction based on exponential moving average mechanisms~\cite{kingma_adam_2017-1}. The update step $t+1$ of task $k$ (Eq.~(\ref{eq:IUS_SGD})) can be then reformulated as:
\begin{equation}
 w_{t+1}^{(k)} =
 \begin{cases}
 w_{t}^{(N)} - \eta_t \cdot \hat{m}\left(g^{(k)}\left(w_{t}^{(N)}, \xi_t \right)\right), \quad k=1
 \\[10pt]
  w_{t+1}^{(k-1)} - \eta_t \cdot \hat{m}\left(g^{(k)}\left(w_{t+1}^{(k-1)}, \xi_t
  \right)\right), \quad \forall \quad k > 1, 
  
 \end{cases}  
 \label{eq:IUS}
\end{equation}
where $\hat{m}$ mixes $g^{(k)}$ with descent directions $g^{({j \neq i})}$ of previous tasks update steps. This implies that using this rule does not allow for fully independent task-specific update steps thanks to $\hat{m}$.
% {\color{red}This individual update step rule, which we denote MTL-IUS, is equivalent to the update step used in partitioning methods~\cite{bragman_stochastic_2019,maninis_attentive_2019-1,pascal_maximum_2020}. What was then section 3.2.1?}

\textbf{Individual Optimizers (MTL-IO).}
We propose to use individual exponential moving averages for each task to allow for fully independent task-specific update steps. We formulate this update rule, using Individual Optimizers (IOs), as:
\begin{equation}
 w_{t+1}^{(k)} =
 \begin{cases}
 w_{t}^{(N)} - \eta_t \cdot \hat{m}^{(k)}\left(g^{(k)}\left(w_{t}^{(N)}, \xi_t\right)\right), \quad k=1
 \\[10pt]
  w_{t+1}^{(k-1)} - \eta_t \cdot \hat{m}^{(k)}\left(g^{(k)}\left(w_{t+1}^{(k-1)}, \xi_t\right)\right), \quad \forall \,\, k > 1 
 \end{cases}  
 \label{eq:IO}
\end{equation}
with $\hat{m}^{(k)}$ the exponential moving average mechanism of task $k$. We denote this alternate and independent optimization strategy as MTL-IO to reflect the properties of its update rule. With MTL-IO, the memory term introduced by $\hat{m}^{(k)}$ only involves previous updates of task $k$. This is equivalent to using one IO per task, which avoids the memory leakage that MTL-IUS incurs into.

\subsection{Mitigating computational costs through task grouping}
\label{subsec:grouping}
Alternate and independent optimization schemes using MTL-IUS or MTL-IO require a training step per task per input sample, i.e $N$ training steps in total. This makes the inference and training time proportional to the number of tasks. %Although this is not suggested by the convex convergence bound (\ref{eq:convergence}), training times also tend to increase with the number of tasks for non-convex settings, according to our results. 
To compensate for the increased computational burden, we propose to equally and randomly distribute the $N$ tasks among a set $\smash{\mathcal{\hat{T}} = \{ \mathcal{\hat{T}}_l \}_{l \in \{1..\hat{N}\}}}$ of $\smash{\hat{N}}$ super-tasks, with $\smash{\hat{N} \in \{1..N\}}$. The set of super-tasks can be optimized with MTL-IUS or MTL-IO, whereas the individual super-task gradients are expressed as the weighted sum of the gradients of their constituting tasks:
\begin{equation}
 w_{t+1}^{(l)} =
 \begin{cases}
 w_{t}^{(\hat{N})} - \eta_t \cdot \hat{m}\left(\sum_{k \in \mathcal{\hat{T}}_l}g^{(k)}(w_{t}^{(\hat{N})}, \xi_t)\right) , \quad l=1
 \\[10pt]
  w_{t+1}^{(l-1)} - \eta_t \cdot \hat{m}\left(\sum_{k \in \mathcal{\hat{T}}_l}g^{(k)}\left(w_{t+1}^{(l-1)}. \xi_t\right)\right), \quad \forall \,\, l > 1 
 \end{cases}  
 \label{eq:groups}
\end{equation}
For the sake of brevity, we only present the formulation adapted for MTL-IUS. Derivation with MTL-IO is straightforward.
One should note that this formulation is equivalent to MTL-IUS (respectively, MTL-IO) when $\hat{N}=T$, and to MTL-SUS when $\hat{N}=1$. Such task grouping thus defines an adjustable compromise between tasks independence and training speed.

% ========================================================================================
% Experimental results
% ========================================================================================

\section{Experiments and results}\label{sec:experiments}
We report four different experiments performed on three widely used datasets for MTL. We first study the generalization performance of MTL-SUS, MTL-IUS and MTL-IO on NYUv2~\cite{silberman_indoor_2012}, with three heterogeneous tasks, and on the seven homogeneous segmentation tasks of Cityscapes~\cite{cordts_cityscapes_2016-2}. We then study how our grouping strategy behaves on the $40$ tasks of Celeb-A~\cite{liu_deep_2015-1}. We finally provide insights about the shared parameter space exploration operated by the studied optimization schemes.

In this study, our models are compared with state-of-the-art multi-task methods, including 
three aggregated loss optimization methods, GradNorm \cite{chen_gradnorm_2018}, MGDA \cite{sener_multi-task_2018-1}, PCGrad \cite{chen_just_2020}, and one partitioning method, Maximum Roaming (MR) \cite{pascal_maximum_2020}.
To ensure fairness of comparison, all baselines have been implemented in the same pipeline, under Pytorch $1.2$, and run on Nvidia Titan-XP GPUs. For every baseline compared on a dataset, we perform a grid search on the learning rate.
All the scripts we used to perform our experiments are available on GitHub.\footnote{\url{https://github.com/lucaspascal/AITSO_MTL}.}. The data used comes from official releases for the three datasets.

\subsection{Scene understanding on NYUv2}
The NYUv2 dataset \cite{silberman_indoor_2012} provides close to $1500$ annotated indoor images extracted from videos captured with the Microsoft Kinect in different buildings. The small number of images and the complexity of the scenes makes it particularly difficult and interesting for multi-task problems. We use it here for one $13$-class semantic segmentation task, one depth estimation task, and one normal estimation task. We judge this multi-task setting as particularly challenging, since it is made of three tasks of different natures. 

All baselines use a U-Net \cite{ronneberger_u-net_2015}, which is known to be particularly efficient for dense labelling tasks. To maximize the proportion of shared weights in the network, all encoders and decoders are shared by all the tasks, while the task-specific prediction heads are made of a single $1\times1$ convolutional layer per task.

% We use the official release of the dataset \cite{silberman_indoor_2012}\footnote{NYUv2 URL}. 
The image size is set to $288 \times 384$ with a batch size of $8$. All models are trained for $500$ epochs, and all performances are reported over $3$ random seeds. Reported metrics are mean Intersection over Union (mIoU) and pixel accuracy for segmentation tasks, absolute and relative error for depth estimation, and mean and median angle error for normal estimation.

To understand the general behavior of the compared optimization strategies MTL-SUS, MTL-IUS and MTL-IO, we first study their generalization performance  w.r.t. the learning rate $\eta$, which has a direct influence on the sensitivity of deep networks to minima in the loss landscape, and therefore on their convergence. 
We report in Fig.~\ref{fig:nyu_loss} the best validation losses achieved by MTL-SUS, MTL-IUS and MTL-IO over a full training, w.r.t. different values of $\eta$. We made sure for every model that the validation loss reached its minimum before the end of the training.

\begin{figure}[h]
 \centering
 \includegraphics[width=1\linewidth]{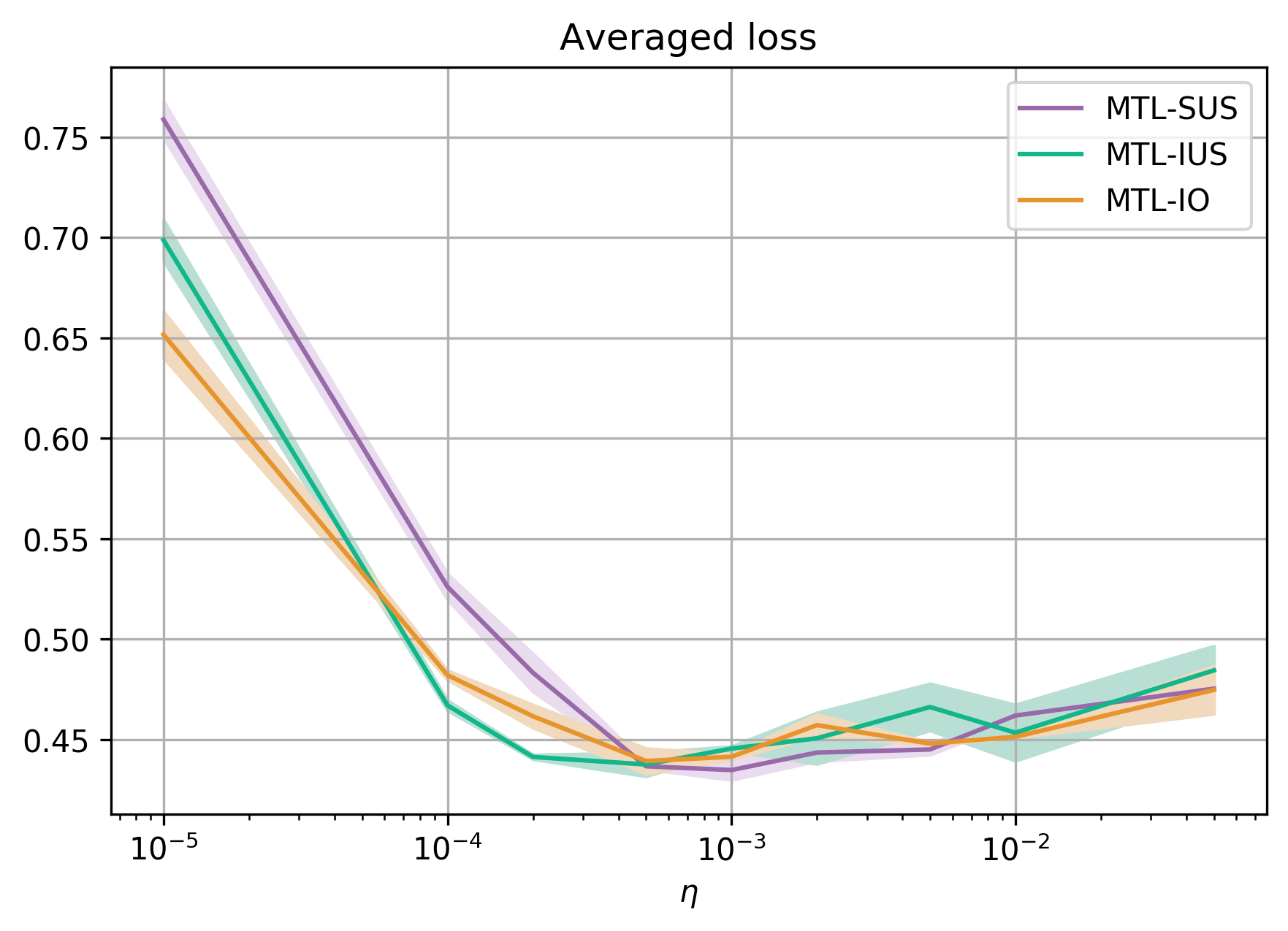}
 \caption{Cityscapes best averaged and task-specific validation losses w.r.t the learning rate $\eta$.}
 \label{fig:city_loss}
\end{figure}

We first observe that MTL-IO is by far the best overall performing method. Above $\eta=5e^{-5}$, it substantially improves the losses of each of the three tasks, which suggests that it is able to consistently reach better loss regions in the shared parameter space. We also observe a smaller difference between MTL-IUS and MTL-SUS. Specifically, MTL-IUS performs better for smaller learning rates, and worse beyond $\eta=2e^{-4}$.
It is also interesting to note the overall distorted aspect of these curves, with high standard deviations, while more "convex" shapes could be expected. This suggests that the loss landscape is particularly complex, compared to that one of other datasets (see Fig.~\ref{fig:city_loss},\ref{fig:celeba_loss}).

We compare the models with the best validation losses for each method with the state-of-the-art baselines (Table~\ref{nyu-table}). While MTL-IUS and MTL-SUS perform similarly, MTL-IO achieves the best results overall on segmentation and depth estimation metrics, and falls close behind MR for normals estimation, making it the best performing multi-task method on this dataset.

\begin{table*}[h]
\small
\centering
\begin{tabular}{l|*{7}{p{13mm}}}
 \toprule
 & \multicolumn{2}{c}{Segmentation} & \multicolumn{2}{c}{Depth estimation} & \multicolumn{2}{c}{Normals estimation} & \\
 \cmidrule(r){2-3} \cmidrule(r){4-5} \cmidrule(r){6-7}
  & mIoU \newline $(\uparrow)$ & Pix. Acc. \newline $(\uparrow)$ & Abs. Err. \newline $(\downarrow)$ & Rel. Err. \newline $(\downarrow)$ & Mean Err. \newline $(\downarrow)$ & Med. Err. \newline $(\downarrow)$ & Rank \newline $(\downarrow)$\\
 \midrule
 MTL-SUS & $24.29 \newline (\pm 0.38)$ & $48.79 \newline (\pm 0.35)$ & $68.48 \newline (\pm 0.28)$ & $52.93 \newline (\pm 0.59)$ & $27.28 \newline (\pm 0.02)$ & $33.70 \newline (\pm 0.03)$ & $4.33$\\
  \midrule
MTL-IUS & $24.93 \newline (\pm 0.33)$ & $48.91 \newline (\pm 0.17)$ & $67.76 \newline (\pm 0.61)$ & $53.25 \newline (\pm 0.77)$ & $27.28 \newline (\pm 0.05)$ & $33.70 \newline (\pm 0.05)$ & $3.88$\\
 \midrule
 MTL-IO & $\mathbf{29.48} \newline \mathbf{(\pm 0.30)}$ & $\mathbf{54.06} \newline \mathbf{(\pm 0.21)}$ & $\mathbf{66.07} \newline \mathbf{(\pm 0.68)}$ & $\mathbf{50.27} \newline \mathbf{(\pm 0.28)}$ & $\mathbf{27.08} \newline \mathbf{(\pm 0.03)}$ & $\mathbf{33.35} \newline \mathbf{(\pm 0.05)}$ & $\mathbf{1.00}$\\
 \midrule
 GradNorm & $28.09 \newline (\pm 0.85)$ & $52.91 \newline (\pm 0.74)$ & $70.26 \newline (\pm 1.4)$ & $52.40 \newline (\pm 1.35)$ & $27.18 \newline (\pm 0.05)$ & $33.44 \newline (\pm 0.07)$ & $3.16$\\
 \midrule
 PCGrad & $24.51 \newline (\pm 0.32)$ & $48.87 \newline (\pm 0.41)$ & $67.89 \newline (\pm 0.67)$ & $52.29 \newline (\pm 0.75)$ & $27.32 \newline (\pm 0.04)$ & $33.65 \newline (\pm 0.08)$ & $3.50$\\
 \midrule
 MR & $28.82 \newline (\pm 0.18)$ & $\mathbf{53.77} \newline \mathbf{(\pm 0.61)}$ & $71.30 \newline (\pm 0.97)$ & $53.37 \newline (\pm 0.97)$ & $\mathbf{27.03} \newline \mathbf{(\pm 0.02)}$ & $\mathbf{33.27} \newline \mathbf{(\pm 0.03)}$ & $2.66$\\
 \bottomrule
\end{tabular}
%  \caption{NYUv2 results. The best results (and statistically equivalent) per column are highlighted in each category. Methods ranks are computed by averaging the rank of the methods on each metric.}
 \caption{NYUv2 results with standard deviations. The best results (and statistically equivalent) per column are highlighted in each category. Methods ranks are computed by averaging the rank of the methods on each metric.}
 \label{nyu-table}
\end{table*}

\subsection{Multi-class segmentation on Cityscapes}
The Cityscapes dataset \cite{cordts_cityscapes_2016-2} provides $5$K fine-grained annotated street-view images captured from a car point of view, in many different German cities. We use here the $7$ semantic segmentation classes as $7$ independent tasks, to observe how the different optimization methods behave with more homogeneous tasks.
We use the same network as for NYUv2, and images are resized to $128 \times 256$ with a batch size of $24$. We conduct a study similar to NYUv2, reported in Fig.~\ref{fig:city_loss}, and a benchmark in Table~\ref{cityscape-table}.

\begin{table}[h]
\centering
\begin{tabular}{l|rr|r}
 \toprule
 & \multicolumn{2}{c}{Segmentation} &\\
 \cmidrule(r){2-3} 
  & \multicolumn{1}{c}{mIoU $(\uparrow)$} & \multicolumn{1}{c}{Pix. Acc. $(\uparrow)$} & \multicolumn{1}{c}{Rank $(\downarrow)$} \\
 
 \midrule
 MTL-SUS & $\mathbf{69.79 \pm 0.40}$ & $\mathbf{90.40 \pm 0.08}$ & $\mathbf{1.0}$\\
 \midrule
 MTL-IUS & $\mathbf{69.74 \pm 0.05}$ & $90.22 \pm 0.13$ & $2.5$\\
 \midrule
 MTL-IO & $69.52 \pm 0.26$ & $\mathbf{90.28 \pm 0.14}$ & $2.5$\\
 \midrule
 GradNorm & $69.00 \pm 0.12$ & $90.17 \pm 0.10$ & $5.0$\\
 \midrule
 PCGrad & $\mathbf{69.73 \pm 0.05}$ & $\mathbf{90.42 \pm 0.06}$ & $\mathbf{1.0}$\\
 \midrule
 MR & $69.15 \pm 0.08$ & $89.79 \pm 0.08$ & $5.5$\\
 \bottomrule
\end{tabular}
%  \caption{Cityscape results. The best results (and statistically equivalent) per column are highlighted. Methods ranks are computed by averaging the rank of the methods on each metric.}
 \caption{Cityscape results with standard deviations. The best results (and statistically equivalent) per column are highlighted. Methods ranks are computed by averaging the rank of the methods on each metric.}
 \label{cityscape-table}
\end{table}

Here we observe smoother curves with smaller variations, which confirms our intuition that this setting is easier to optimize. The three optimization methods report very similar results. While the best overall validation loss is reached by MTL-SUS, MTL-IUS and MTL-IO are consistently better in the lower learning rates, suggesting that they are more robust to local minima in the loss landscape. As one could expect given Fig.~\ref{fig:city_loss}, all the methods, including the  state-of-the art benchmarks, present a very close performance. It seems like in a simple setting with a well calibrated learning rate, the influence of a particular optimization method is greatly reduced.

\subsection{Multi-attribute segmentation on Celeb-A}
The CelebA dataset~\cite{liu_deep_2015-1} contains $200$K face images of $10$K different celebrities. Annotations are provided for $40$ different facial attributes. 
In this experiment, we study the proposed random grouping strategy with a large number of tasks, to establish to which extent it can reduce the computational overhead of MTL-IUS and MTL-IO without affecting their performance.
We consider each of the $40$ provided facial attributes as independent facial attribute detection task, and operate different groupings on them, $\hat{N}=\{2, 4, 8, 20, 40\}$.

\begin{figure*}[t]
 \centering
 \includegraphics[width=0.8\linewidth]{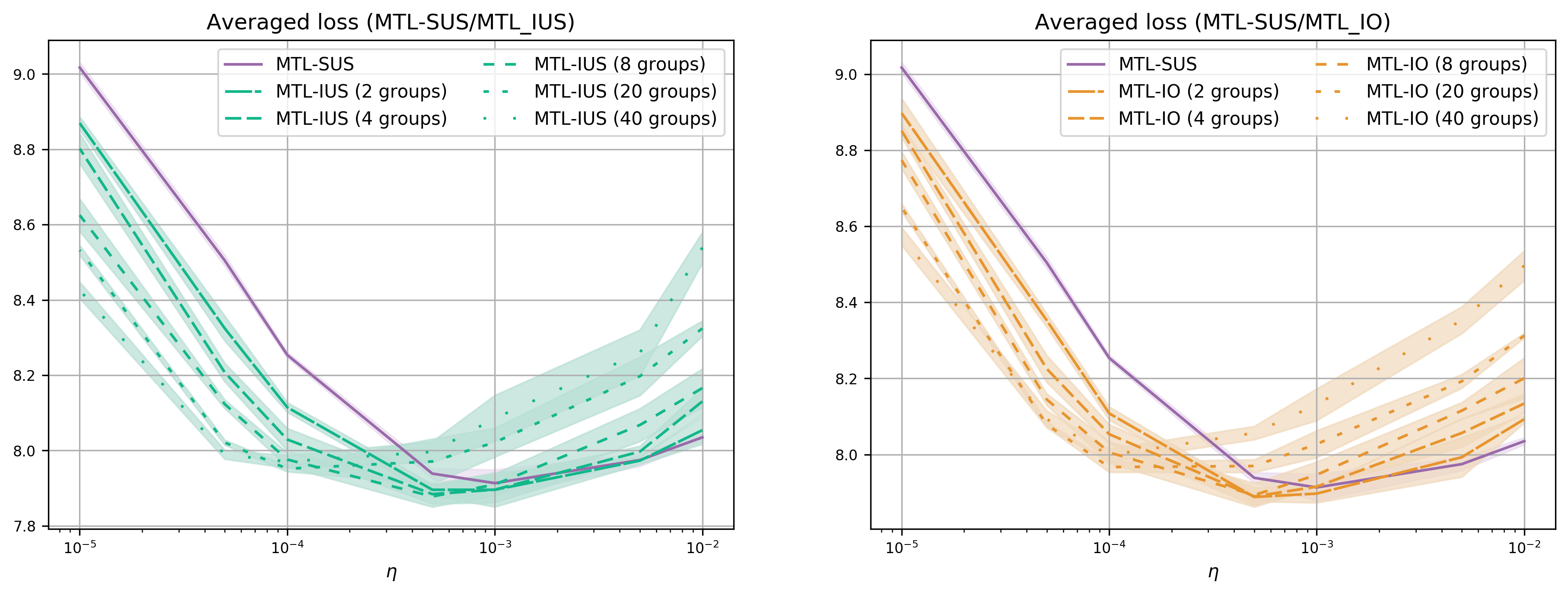}
 \caption[Man a woman]{Celeb-A best averaged validation losses w.r.t the learning rate $\eta$ and different numbers of task groups. \textbf{(Left)} MTL-IUS models compared to MTL-SUS. \textbf{(Right)} MTL-IO models compared to MTL-SUS.}
 \label{fig:celeba_loss}
% \end{wrapfigure}
\end{figure*}

We use a fully shallow $9$-layer convolutional network similar to \cite{chen_just_2020}, with a task-specific fully-connected prediction layer for each task (or task group). The image size is set to $64\times64$, with a batch size of $256$. All models are trained over $15$ epochs. We report the average classification error, along with F1-score, since the different attributes are not balanced. Reported results are averaged over $3$ seeds, with new task groups are sampled for every seed.
Figure~\ref{fig:celeba_loss} shows the best achieved validation losses over training  w.r.t. the learning rate and for different number of super-tasks $\smash{\hat{N}}$. Table ~\ref{celeba_table} summarizes the benchmark result. 

For a certain number of super-tasks, i.e. $\smash{\hat{N} \leq 8}$, MTL-IUS and MTL-IO are able to reach an improvement compared to MTL-SUS. Then, the performance decreases up to $\smash{\hat{N} = 40}$ (i.e. no grouping). This suggests that the use of a grouping strategy is beneficial both in terms of performance and computational efficiency when dealing with a large number of tasks. The latter is further confirmed in Table~\ref{celeba_table}, where the best grouping models of MTL-IUS and MTL-IO also outperform other state-of-the art methods on both average error and F-score metrics.

%
% {\color{red}We observe that while $40$ tasks seem too much to handle for MTL-IUS and MTL-IO, they are able to reach some improvement compared to MTL-SUS when using the random grouping strategy for a number of super-tasks $\hat{N} \leq 8$, and we are thus able to reach a good compromise between performance and computational overhead. NOT CLEAR. Split into multiple sentences} 

\begin{table*}[h]
\centering
\begin{tabular}{l|rrr}
 \toprule
 & Avg. error $(\downarrow)$ & F1-score $(\uparrow)$ & Rank $(\downarrow)$\\
 \midrule
 MTL-SUS & $9.20 \pm 0.02$ & $68.66 \pm 0.24$ & $6.0$\\
 \midrule
 MTL-IUS ($2$ groups) & $9.16 \pm 0.03$ & $68.69 \pm 0.21$ & $-$\\
 MTL-IUS ($4$ groups) & $9.13 \pm 0.02$ & $69.01 \pm 0.27$ & $-$\\
 MTL-IUS ($8$ groups) & $\mathbf{9.10 \pm 0.03}$ & $68.93 \pm 0.38$ & $3.0$\\
 MTL-IUS ($20$ groups) & $9.20 \pm 0.05$ & $68.82 \pm 0.11$ & $-$\\
 MTL-IUS ($40$ groups) & $9.27 \pm 0.05$ & $68.97 \pm 0.17$ & $-$\\
 \midrule
 MTL-IO ($2$ groups) & $9.15 \pm 0.02$ & $68.85 \pm 0.33$ & $-$\\
 MTL-IO ($4$ groups) & $\mathbf{9.11 \pm 0.02}$ & $\mathbf{69.48 \pm 0.15}$ & $\mathbf{1.0}$\\
 MTL-IO ($8$ groups) & $9.15 \pm 0.01$ & $68.55 \pm 0.09$ & $-$\\
 MTL-IO ($20$ groups) & $9.23 \pm 0.02$ & $68.61 \pm 0.31$ & $-$\\
 MTL-IO ($40$ groups) & $9.24 \pm 0.00$ & $68.57 \pm 0.04$ & $-$\\
 \midrule
 GradNorm & $9.16 \pm 0.03$ & $69.16 \pm 0.19$ & $4.0$\\
 PCGrad & $9.16 \pm 0.04$ & $\mathbf{69.26 \pm 0.30}$ & $2.5$\\
 MGDA-UB & $9.42 \pm 0.03$ & $67.68 \pm 0.12$ & $7.0$\\
 MR & $\mathbf{9.12 \pm 0.00}$ & $\mathbf{69.21 \pm 0.31}$ & $\mathbf{1.0}$\\
 \bottomrule
\end{tabular}
%  \caption{Celeb-A results. The best results (and statistically equivalent) per column are highlighted. In parenthesis are reported the different number of task groups for MTL-IUS and IO. We report the best ($4$ and $8$) and worst ($40$) performing numbers of task groups for both the strategies.}
 \caption{Celeb-A results with standard deviations. The best results (and statistically equivalent) per column are highlighted. In parenthesis are reported the different number of task groups for MTL-IUS and IO. The best ($4$ and $8$) and worst ($40$) performing numbers of task groups are reported for both the strategies.}
 \label{celeba_table}
\end{table*}

More generally, we observe a progressive evolution of the curves w.r.t. the different values of $\hat{N}$: as with Cityscapes, the less we group tasks together, the better is the optimization of IUS and IO in the low learning rates, while it is the opposite in the high learning rates. This behavior suggests that these methods introduce more noise in the optimization, making them less sensitive to local minima, and pushing for a larger exploration of the shared parameter space.

% \begin{wrapfigure}{r}{\textwidth}

\subsection{Covered distance}
We provide insights about how the studied optimization strategies explore the shared parameter space, by measuring the distances covered in this space during training. Using the Frobenius norm, we measure the shortest path from the network's initialization to the loss minimum (shortest) and the total covered distance (total), which is the sum of the distances covered at each update step. We also report the total-to-shortest-distance ratio, which gives an insight on how much the surrounding space has been explored for a given shortest path.

Table~\ref{tab:traveled_distance} summarizes the obtained measures during the optimization of MTL-SUS, MTL-IUS and MTL-IO over all datasets. All three baselines start from the same parameter initialization (i.e. same point in the parameter space) and use the same learning rate (the one providing the best performances overall). The measurements are performed at the same point in time (the closest from the validation loss minimum). The results are averaged over 3 different seeds. As a reminder, the aggregated multi-task loss uses uniform weights of $1$ for every task, so that the gradient scaling is the same for the three optimization strategies.

\begin{table*}[t]
 \centering
 \begin{tabular}{l|l|rrr}
 \toprule
 Dataset & Model & Total covered dist. & Shortest path & Ratio \\
 \midrule
 \multirow{3}{4em}{NYUv2} & SUS & $879.33 \pm 18.22$ & $38.77 \pm 0.53$ & $22.68 \pm 0.17$ \\
 & IUS & $2597.74 \pm 49.59$ & $67.44 \pm 1.02$ & $38.52 \pm 0.24$ \\
 & IO & $2769.52 \pm 40.20$ & $71.06 \pm 0.80$ & $38.97 \pm 0.22$ \\
 \midrule
 \multirow{3}{4em}{Cityscapes} & SUS & $909.38 \pm 10.02$ & $74.38 \pm 0.80$ & $12.23 \pm 0.01$ \\
 & IUS & $8102.39 \pm 13.16$ & $262.68 \pm 0.48$ & $30.85 \pm 0.02$ \\
 & IO & $8438.78 \pm 105.47$ & $277.29 \pm 5.69$ & $30.44 \pm 0.35$ \\
 \midrule
 \multirow{11}{4em}{Celeb-A} & SUS & $1398.01 \pm 3.54$ & $99.66 \pm 0.22$ & $14.03 \pm 0.00$ \\
 & IUS (4 groups) & $5636.18 \pm 19.59$ & $223.83 \pm 0.69$ & $25.18 \pm 0.02$ \\
 & IO (4 groups) & $6560.56 \pm 45.21$ & $242.55 \pm 1.80$ & $27.05 \pm 0.12$ \\
 & IUS (8 groups) & $14955.87 \pm 15.55$ & $387.13 \pm 0.48$ & $38.63 \pm 0.01$ \\
 & IO (8 groups) & $13992.50 \pm 268.21$ & $369.67 \pm 5.39$ & $37.85 \pm 0.17$ \\
 & IUS (40 groups) & $65376.36 \pm 164.54$ & $837.32 \pm 1.91$ & $78.08 \pm 0.08$ \\
 & IO (40 groups) & $59627.39 \pm 746.35$ & $764.42 \pm 9.01$ & $78.00 \pm 0.07$ \\
 \bottomrule
 \end{tabular}
 \caption{Total covered and shortest path distances in the shared parameter space from a same initialization point.}
 \label{tab:traveled_distance}
\end{table*}

We observe that both MTL-IUS and MTL-IO travel more, both in terms of total covered and shortest path distances. The distances increase globally with the number of tasks (or task groups).
The increase in the shortest path distance indicates that these methods are able to discover more distant loss regions compared to MTL-SUS.
These observations correlate with the findings from our previous experiments, suggesting that some of these more distant regions can provide better generalization performance.

Most interestingly, MTL-IUS and MTL-IO present a higher total-to-shortest path distance ratio compared to MTL-SUS, which means that between two locations in the shared parameter space, MTL-IUS and MTL-IO follow a much more oscillating trajectory.
This observation suggests that optimizing tasks-specific objective functions, instead of an aggregated multi-task objective one, introduces more stochasticity to the optimization, which is usually beneficial in deep learning.

% Surprisingly, MTL-IUS and MTL-IO obtain almost similar values, while we could expect MTL-IO to cover more distance, since it applies fully independent task-specific update steps. The very low standard deviations of these observations (run on $3$ seeds) are also intriguing, since the results are obtained from different random seeds. 
% \color{red} and why is the reason for this? }

% ========================================================================================
% Conclusion
% ========================================================================================
\section{Conclusion}\label{sec:conclusion}

We have studied the \textit{alternate and independent} optimization of the task-specific objective functions (MTL-IUS) used in partitioning methods, and compared it to the optimization of the more standard aggregated objective function (MTL-SUS), which is widely adopted in the literature. Taking inspiration from Federated Learning works, we first demonstrated that the former has similar convergence properties as the latter when dealing with convex objective functions. We then showed that the existing partitioning methods do not operate truly independent update steps due to the momentum mechanisms included in state-of-the-art optimizers. We thus have formulate a novel strategy, using an independent optimizer per task (MTL-IO), that favors task independence.
To account for the computational overhead of these strategies, we have also proposed a random grouping strategy, that strikes a balance between computational efficiency,  which has the merits of potentially reducing the carbon footprint of the proposed methods, and the benefits of a task-independent optimization strategy.

% {\color{red} Refactor conclusions to be more aligned with contributions in intro, recall the alternate and independent bla bla. In the conclusions, mention the analogy with federated learning.}

Our experimental results over three datasets show both MTL-IUS and MTL-IO achieve an overall better generalization performance w.r.t standard aggregated objective optimization (MTL-SUS) and state-of-the-art MTL baselines using an aggregated objective function. 
In particular, MTL-IO shows important improvements in settings where tasks are of a very different nature. Our results also confirm that the proposed random grouping strategy applied to MTL-IUS and MTL-IO is beneficial both in terms of performance and efficiency, when dealing with a great number of tasks. Finally, we showed that MTL-IUS and IO allow parameters updates to travel a longer distance, and ensure a more thorough exploration of the shared parameter space.

% We finally showed that we are able to efficiently reduce the computational overhead of these methods with a simple random grouping strategy.

While the optimization of an aggregated multi-task objective function is widely adopted in the literature, our investigations suggest that its usage should be questioned, %when applied to deep networks, 
despite the computational benefits it provides. 
We hope this work will inspire further studies in this direction, as many existing optimization methods of the aggregated multi-task objective function could be adapted to task-specific update steps.

% Further studies should be carried on to compare it to an alternate and independent optimization of the task-specific objective functions. We encourage research in this direction, as many existing optimization methods of the aggregated multi-task objective function could be adapted to task-specific update steps.
%------------------------------------------------------------------------

\newpage
\bibliographystyle{plain}
\bibliography{biblio}

%%%%%%%%%%%%%%%%%%%%%%%%%%%%%%%%%%%%%%%%%%%%%%%%%%%%%%%%%%%%

\newpage 

\section{Proof of Theorem~\ref{theorem:convergence}}
\label{app:proof}
In this section we prove the convergence of MTL-IUS for stochastic gradient descent when dealing with convex objective functions (Theorem~\ref{theorem:convergence}). We take inspiration from the proof of convergence of the FedAvg algorithm~\cite{li2020convergence} for partial device participation in federated learning. %, and essentially follows the same structure.

\subsection{Problem setting and notations}
\label{app:th_notations}
For the ease of notation, in the following we drop the double indexing of $w$ (tasks and iterations) used in equation~\ref{eq:IUS_SGD}. We thus rewrite the IUS update rule (i.e. equation~\ref{eq:IUS_SGD}) of the weights at update step $t+1$ as:
\begin{equation*}
 \begin{cases}
  v_{t+1}^k = w_t - \eta_t \nabla F^{(k)}(w_t, \xi_t^k), \\[10pt]
  w_{t+1} = v_{t+1}^{s_t},
 \end{cases}  
\end{equation*}
where $v_{t+1}^k$ is the SGD from $w_t$ with respect to task $k$ objective function, 
and $w_{t+1}$ only follows the SGD of one task $s_t$ (while others are not considered), with $s_t$ selected in $\{1,...,N\}$, for every step $t$.
We prove here the theorem for $s_t$ randomly selected at uniform at each iteration $t$, which also proves it for our case, since the proof only assumes a sampling probability of $p_k = \frac{1}{N}$ for any task $k$.

We define $\bar{v}_t = \frac{1}{N} \sum_{k=1}^N v_t^k$ the averaged SGD at step $t+1$, $\bar{g}_t = \frac{1}{N} \sum_{k=1}^N \nabla F^{(k)}(w_t)$ the averaged gradient and $g_t = \frac{1}{N} \sum_{k=1}^N \nabla F^{(k)}(w_t, \xi_t^k)$ the averaged stochastic gradient. We have $\bar{v}_{t+1} = w_t - \eta_t g_t$ and $\esp g_t = \bar{g}_t$.

\subsection{Key lemmas}
We first state two lemmas used in the theorem's proof. We leave their proof for Appendix \ref{app:lemma_proofs}.

\begin{lemma}
\label{lem:lemma_1}
Under assumptions~\ref{ass:smooth} and~\ref{ass:convex}, if $\eta_t \leq \frac{1}{L}$, we have:
\begin{align}
 \mathbb{E}||\overline{v}_{t+1} - w^*||^2 \leq & (1-\mu \eta_t) \mathbb{E}||w_t - w^*||^2 \nonumber \\
 & + 2L\eta_t^2\Gamma + \frac{\eta_t^2}{N^2} \sum_{k=1}^N \sigma_k^2
\end{align}
\end{lemma}

\begin{lemma}
\label{lem:lemma_2} For $\eta _t$ non-increasing such as $\eta_t \leq 2\eta_{t+1}, \forall t \geq 0$, we have:
\begin{equation}
 \mathbb{E}_{s_t}||\overline{v}_{t+1} - w_{t+1}||^2 \leq \eta_t^2 G^2
\end{equation}
\end{lemma}

\subsection{Theorem proof}
\label{app:th_proof}
\begin{align*}
 ||w_{t+1} - w^*||^2 = & \underbrace{||w_{t+1} - \overline{v}_{t+1}||^2}_{A_1} + \underbrace{||\overline{v}_{t+1} - w^*||^2}_{A_2} \\
 & + \underbrace{2<w_{t+1} - \overline{v}_{t+1}, \overline{v}_{t+1} - w^*>}_{A_3}
\end{align*}

For $A_3$, we have
\begin{equation*}
 \esp_{s_t} (w_{t+1}) = \esp_{s_t} (v_{t+1}^{s_t}) = \frac{1}{N} \sum_{k=1}^N v_{t+1}^k = \bar{v}_{t+1}
\end{equation*}

which proves $\mathbb{E}_{s_{t+1}}(A_3) = 0$, where $\esp_{s_t}$ accounts for the expectation taken over the random sampling of $s_t$.

We bound $A_1$ with Lemma 2 and $A_2$ with Lemma 1. We get:
\begin{align*}
 \mathbb{E}||w_{t+1} - w^*||^2 \leq & (1-\mu \eta_t) \mathbb{E}||w_t - w^*||^2 \\
 & + \frac{\eta_t^2}{N^2}\sum_{k=1}^N \sigma_k^2 + 2L\eta_t^2\Gamma + \eta_t^2 G^2
\end{align*}
Let $\Delta_t = \esp\norms{w_{t+1} - w^*}$ and $B = \frac{1}{N^2}\sum_{k=1}^N \sigma_k^2 + 2L\Gamma + G^2$. We obtain:

\begin{equation*}
 \Delta_{t+1} \leq (1-\mu\eta_t)\Delta_t + \eta_t^2B
\end{equation*}

% We then show by induction that $\mathbb{E}||w_{t+1} - w^*||^2 \leq \frac{v}{\gamma + t}$ with $v = \max \{ \frac{\beta^2B}{\beta\mu - 1}, (\gamma+1)||w_1 - w^*||^2 \}$, $B=2L\Gamma$, $C=G^2$, $\beta > \frac{1}{\mu}$, $\gamma > 0$ such that $\eta_1 \leq \min \{ \frac{1}{\mu}, \frac{1}{4L} \} = \frac{1}{4L}$, $\eta_t \leq 2\eta_{t+1}$:

We show by induction that for $\beta > \frac{1}{\mu}$ and $\gamma > 0$ verifying $\eta_1 \leq \frac{1}{L}$ and $\eta_t \leq 2\eta_{t+1}$, we have $\Delta_t \leq \frac{v}{\gamma + t}$ with $v = \max \{ \frac{\beta^2B}{\beta\mu - 1}, (\gamma+1)\Delta_1 \}$:

The affirmation directly holds for $t=1$. Then, suppose it holds for $t$, we have:
\begin{align*}
 \Delta_{t+1} & \leq (1-\mu\eta_t)\Delta_t + \eta_t^2B \\
 & \leq \left(1-\frac{\beta\mu}{t+\gamma}\right)\frac{v}{t+\gamma} + \frac{\beta^2B}{(t+\gamma)^2} \\
 & = \frac{t+\gamma-1}{(t+\gamma)^2}v + \left[\frac{\beta^2B}{(t+\gamma)^2} - \frac{\beta\mu - 1}{(t+\gamma)^2}v\right] \\
 & \leq \frac{t+\gamma-1}{(t+\gamma)^2}v \\
 & \leq \frac{t+\gamma-1}{(t+\gamma+1)(t+\gamma-1)}v = \frac{v}{(t+1)+\gamma} \\
\end{align*}

where the second inequality comes from $v \geq \frac{\beta^2B}{\beta\mu-1}$. This proves that the affirmation holds for $t+1$.

With the convexity of $F$, we obtain:
\begin{multline*}
 \mathbb{E}[F(w_t)] - F^* \leq \frac{L}{2(\gamma + t)} \left( \frac{\beta^2B}{\beta\mu - 1} + (\gamma+1)||w_1 - w^*||^2 \right) \\
\end{multline*}

which proves theorem~\ref{theorem:convergence} for $\beta=\frac{2}{\mu}$, and $\gamma \geq 2\frac{L}{\mu}-1$, to ensure $\eta_1 \leq \frac{1}{L}$.

\subsection{Proof of key lemmas}
\label{app:lemma_proofs}
\textbf{Lemma 1}
We assume assumptions~\ref{ass:smooth} and~\ref{ass:convex} hold, and want to find an upper bound for $\mathbb{E}||\overline{v}_{t+1} - w^*||^2$:
\begin{align}
 \norms{\bar{v}_{t+1} - w^*} & = \norms{w_t - \eta_t g_t - w^* - \eta_t \bar{g}_t + \eta_t \bar{g}_t} \nonumber \\
 & = \underbrace{\norms{w_t - w^* - \eta_t \bar{g}_t}}_{A_1} \nonumber \\
 & \quad + \underbrace{2\eta_t \scal{w_t - w^* - \eta_t \bar{g}_t}{\bar{g}_t - g_t}}_{A_2} \nonumber \\
 & \quad + \eta_t^2\norms{g_t - \bar{g}_t}
 \label{proof_main_eq}
\end{align}

We have $\esp [A_2] = 0$, so we focus on $A_1$:

\begin{align*}
 A_1 & = \norms{w_t - w^* - \eta_t \bar{g}_t} \\
 & = \norms{w_t - w^*} \underbrace{- 2\eta_t \scal{w_t - w^*}{\bar{g}_t}}_{B_1} + \underbrace{\eta_t^2 \norms{\bar{g}_t}}_{B_2} \\
\end{align*}

By convexity of $\norms{.}$ and $L$-smoothness of $F^{(k)}(.)$, we get:
\begin{align*}
 B_2 & = \eta_t^2 \norms{\bar{g}_t} \\
 & \leq \frac{\eta_t^2}{N}\sum_{k=1}^N \norms{\nabla F^{(k)}(w_t)} \\
 & \leq 2L \frac{\eta_t^2}{N}\sum_{k=1}^N (F^{(k)}(w_t) - {F^{(k)}}^*) \\
\end{align*}

Then:
\begin{align*}
 B_1 & = - 2\eta_t \scal{w_t - w^*}{\bar{g}_t} \\
 & = - 2 \frac{\eta_t}{N}\sum_{k=1}^N \scal{w_t - w^*}{\nabla F^{(k)}(w_t)} \\
\end{align*}

The $\mu$-strong convexity of $F^{(k)}(.)$ gives:
\begin{align*}
 -\scal{w_t - w^*}{\nabla F^{(k)}(w_t)} \leq & -(F^{(k)}(w_t) - F^{(k)}(w^*)) \\
 & - \frac{\mu}{2}\norms{w_t - w^*} \\
\end{align*}

we replace in $A_1$:
\begin{align*}
 A_1 & = \norms{w_t - w^* - \eta_t \bar{g}_t} \\
 & \leq \norms{w_t - w^*} + 2L \frac{\eta_t^2}{N}\sum_{k=1}^N (F^{(k)}(w_t) - {F^{(k)}}^*) \\
 & - 2 \frac{\eta_t}{N}\sum_{k=1}^N \left( (F^{(k)}(w_t) - F^{(k)}(w^*)) + \frac{\mu}{2}\norms{w_t - w^*} \right) \\
 & = (1-\mu \eta_t) \norms{w_t - w^*} \\
%  & + \underbrace{2L \frac{\eta_t^2}{N}\sum_{k=1}^N (F^{(k)}(w_t) - {F^{(k)}}^*) - 2\frac{\eta_t}{N}\sum_{k=1}^N \left( F^{(k)}(w_t) - F^{(k)}(w^*) \right)}_C \\
 & + 2L \frac{\eta_t^2}{N}\sum_{k=1}^N (F^{(k)}(w_t) - {F^{(k)}}^*) \\
 & - 2\frac{\eta_t}{N}\sum_{k=1}^N \left( F^{(k)}(w_t) - F^{(k)}(w^*) \right) \\
\end{align*}

We note:
\begin{align*}
    C & = 2L \frac{\eta_t^2}{N}\sum_{k=1}^N (F^{(k)}(w_t) - {F^{(k)}}^*) \\
      & \quad - 2\frac{\eta_t}{N}\sum_{k=1}^N \left( F^{(k)}(w_t) - F^{(k)}(w^*) \right)
\end{align*}

Let $\gamma_t = 2\eta_t (1-L\eta_t)$. Since $\eta_t \leq \frac{1}{L}$, we obtain $\gamma_t \geq 0$. Then:
\begin{align*}
 C & = 2L  \frac{\eta_t^2}{N}\sum_{k=1}^N (F^{(k)}(w_t) - {F^{(k)}}^*) \\
 & \quad - 2 \frac{\eta_t}{N}\sum_{k=1}^N \left( F^{(k)}(w_t) - F^{(k)}(w^*) \right) \\
   & = 2L  \frac{\eta_t^2}{N}\sum_{k=1}^N (F^{(k)}(w_t) - {F^{(k)}}^*) \\
   & \quad - 2 \frac{\eta_t}{N}\sum_{k=1}^N \left( F^{(k)}(w_t) + {F^{(k)}}^* - {F^{(k)}}^* - F^{(k)}(w^*) \right) \\
   & = \left[ \frac{1}{N}\sum_{k=1}^N (F^{(k)}(w_t) - {F^{(k)}}^*) \right] (2L\eta_t^2 - 2\eta_t) \\
   & \quad + 2\frac{\eta_t}{N}\sum_{k=1}^N (F^{(k)}(w^*) - {F^{(k)}}^*) \\
   & = - \frac{\gamma_t}{N}\sum_{k=1}^N (F^{(k)}(w_t) - {F^{(k)}}^*) \\
   & + 2 \frac{\eta_t}{N}\sum_{k=1}^N  (F^{(k)}(w^*) - {F^{(k)}}^*) \\
   & = - \frac{\gamma_t}{N}\sum_{k=1}^N (F^{(k)}(w_t) - F^* + F^* - {F^{(k)}}^*) \\
   & \quad + 2 \frac{\eta_t}{N}\sum_{k=1}^N  (F^* - {F^{(k)}}^*) \\
   & = - \frac{\gamma_t}{N}\sum_{k=1}^N (F^{(k)}(w_t) - F^*) +  \frac{2\eta_t - \gamma_t}{N}\sum_{k=1}^N  (F^* - {F^{(k)}}^*) \\
   & = - \frac{\gamma_t}{N}\sum_{k=1}^N (F^{(k)}(w_t) - F^*) + 2L \frac{\eta_t^2}{N}\sum_{k=1}^N  (F^* - {F^{(k)}}^*) \\
   & = - \frac{\gamma_t}{N}\sum_{k=1}^N (F^{(k)}(w_t) - F^*) + 2L\eta_t^2 \Gamma \\
   & = -\gamma_t(F(w_t) - F^*) + 2L\eta_t^2 \Gamma \leq 2L\eta_t^2 \Gamma \\
\end{align*}

Since $(F(w_t) - F^*) \geq 0$ and $\gamma_t \geq 0$. We replace in equation~\ref{proof_main_eq}:

\begin{align*}
 \norms{\bar{v}_{t+1} - w^*} & \leq (1-\mu\eta_t)\norms{w_t - w^*} \\
 & \quad + 2L\eta_t^2 \Gamma + \eta_t^2\norms{g_t - \bar{g}_t}\\
 \esp \norms{\bar{v}_{t+1} - w^*} & \leq (1-\mu\eta_t)\esp\norms{w_t - w^*} \\
 & \quad + 2L\eta_t^2 \Gamma + \eta_t^2 \esp \norms{g_t - \bar{g}_t}\\
\end{align*}

Then, Assumption 3 gives us:
\begin{align*}
 \esp \norms{g_t - \bar{g}_t} & = \esp \norms{\frac{1}{N}\sum_{k=1}^N (\nabla F^{(k)}(w_t, \xi_t^k) - \nabla F^{(k)}(w_t))} \\
 & = \frac{1}{N^2}\sum_{k=1}^N \esp \norms{\nabla F^{(k)}(w_t, \xi_t^k) - \nabla F^{(k)}(w_t)} \\
 & \leq \frac{1}{N^2}\sum_{k=1}^N \sigma_k^2 \\
\end{align*}

We finally obtain:
\begin{align*}
 \esp \norms{\bar{v}_{t+1} - w^*} & \leq (1-\mu\eta_t)\esp\norms{w_t - w^*} \\
 & \quad + 2L\eta_t^2 \Gamma + \frac{\eta_t^2}{N^2}\sum_{k=1}^N \sigma_k^2\\
\end{align*}

\textbf{Lemma 2}

We now want to find an upper bound for $\esp_{s_t} \norms{\bar{v}_{t+1} - w_{t+1}}$, with $\esp_{s_t}$ the expected value taken over $s_t$ values:
\begin{align*}
 \esp_{s_t} \norms{\bar{v}_{t+1} - w_{t+1}} & = \esp_{s_t} \norms{v_{t+1}^{s_t} - \bar{v}_{t+1}} \\
 & = \frac{1}{N}\sum_{k=1}^N \norms{v_{t+1}^k - \bar{v}_{t+1}} \\
 & = \frac{1}{N} \sum_{k=1}^N \norms{(v_{t+1}^k - w_t) - (\bar{v}_{t+1} - w_t)} \\
 & = \esp \norms{(v_{t+1}^k - w_t) - \esp(v_{t+1}^k - w_t)} \\
\end{align*}
 
Since $\esp \norms{x - \esp x} \leq \esp\norms{x}$, we obtain:
\begin{align*}
 \esp_{s_t} \norms{\bar{v}_{t+1} - w_{t+1}} & \leq \esp \norms{v_{t+1}^k - w_t} \\
 & = \frac{1}{N} \sum_{k=1}^N \norms{v_{t+1}^k - w_t}\\
 & \leq \frac{1}{N} \sum_{k=1}^N \esp \norms{v_{t+1}^k - w_t}\\
 & \leq \frac{1}{N} \sum_{k=1}^N \esp \norms{\eta_t \nabla F^{(k)}(w_t, \xi_t^k)}\\
 & \leq \frac{1}{N} \sum_{k=1}^N \eta_t^2 G^2 = \eta_t^2 G^2\\
\end{align*}
which proves the Lemma.

\end{document}